\documentclass[letterpaper, 10 pt, conference]{ieeeconf}
\usepackage[utf8]{inputenc}

\usepackage{adjustbox}
\usepackage{amsmath} 
\DeclareMathOperator*{\argmax}{arg\,max}

\usepackage[version=4]{mhchem}
\usepackage{siunitx}
\usepackage{longtable,tabularx}
\usepackage{mathrsfs}

\usepackage{hyperref}       % hyperlinks
\usepackage{url}            % simple URL typesetting
\usepackage{booktabs}       % professional-quality tables
\usepackage{amsfonts}       % blackboard math symbols
\usepackage{nicefrac}       % compact symbols for 1/2, etc.
\usepackage{microtype}      % microtypography
\usepackage{color}
\usepackage{csquotes}

\usepackage{graphicx}
\graphicspath{ {images/} }
\usepackage{scrextend}
\usepackage{booktabs}
\usepackage{float}
\usepackage{listings}

\usepackage{pgf}
\usepackage{tikz}
\usetikzlibrary{arrows,automata}
\usetikzlibrary{math} 
\usetikzlibrary{shapes}
\usetikzlibrary{positioning}
\usetikzlibrary{arrows.meta} 
\usetikzlibrary{decorations}
\usetikzlibrary{decorations.pathmorphing}
\usepackage{tkz-euclide}
\usetkzobj{all}

\usepackage{subcaption}

\usepackage{algorithm}
\usepackage{algorithmicx}
\usepackage[noend]{algpseudocode}

\usepackage{relsize}

\usepackage{color, colortbl}
\definecolor{Gray}{rgb}{0.88,0.88,0.88}

\tikzstyle{block} = [draw,fill=blue!20,minimum size=2em]
\tikzstyle{branch}=[fill,shape=circle,minimum size=3pt,inner sep=0pt]

\newcommand{\fastmdp}{\texttt{FastMDP} }

\makeatletter
\newcommand\fs@betterruled{%
  \def\@fs@cfont{\bfseries}\let\@fs@capt\floatc@ruled
  \def\@fs@pre{\vspace*{5pt}\hrule height.8pt depth0pt \kern2pt}%
  \def\@fs@post{\kern2pt\hrule\relax}%
  \def\@fs@mid{\kern2pt\hrule\kern2pt}%
  \let\@fs@iftopcapt\iftrue}
\floatstyle{betterruled}
\restylefloat{algorithm}
\makeatother

\title{An Efficient Algorithm for Multiple-Pursuer-Multiple-Evader Pursuit/Evasion Game}

\author{
  Joshua R. Bertram    \quad   Peng Wei \\
  Iowa State University\\
  Ames, IA 50011 \\
  \texttt{\{bertram1, pwei\}@iastate.edu} \\
}

\begin{document}

\maketitle

\begin{abstract}
    We present a method for pursuit/evasion that is highly efficient and and scales to large teams of aircraft.  The underlying algorithm is an efficient algorithm for solving Markov Decision Processes (MDPs) that supports fully continuous state spaces.  We demonstrate the algorithm in a team pursuit/evasion setting in a 3D environment using a pseudo-6DOF model and study performance by varying sizes of team members.  We show that as the number of aircraft in the simulation grows, computational performance remains efficient and is suitable for real-time systems.  We also define probability-to-win and survivability metrics that describe the teams' performance over multiple trials, and show that the algorithm performs consistently.  We provide numerical results showing control inputs for a typical 1v1 encounter and provide videos for 1v1, 2v2, 3v3, 4v4, and 10v10 contests to demonstrate the ability of the algorithm to adapt seamlessly to complex environments.
\end{abstract}

\section{Introduction}

Pursuit/evasion games pit two opponents against each other such that the pursuer must capture the evader.  Within the aerospace community, pursuit/evasion of aircraft has long been of interest and is seeing a resurgence of interest due to a growing capability and acceptance of autonomous unmanned aircraft.
Additionally, pursuit/evasion games are interesting in that they pose scalability challenges especially to UAV swarm applications.  Problem formulations which lead to efficient and effective pursuit/evasion for 1 versus 1 (1v1) contests do not always allow efficient formulation with larger contests with multiple members per team (e.g., 2v2, 10v10).  For problem formulations and algorithms that can support larger teams, it may be possible to solve the problem offline, but it may be exponentially harder and challenging in an online manner.

\begin{figure}[tbp]
\centering
  \includegraphics[width=.95\columnwidth]{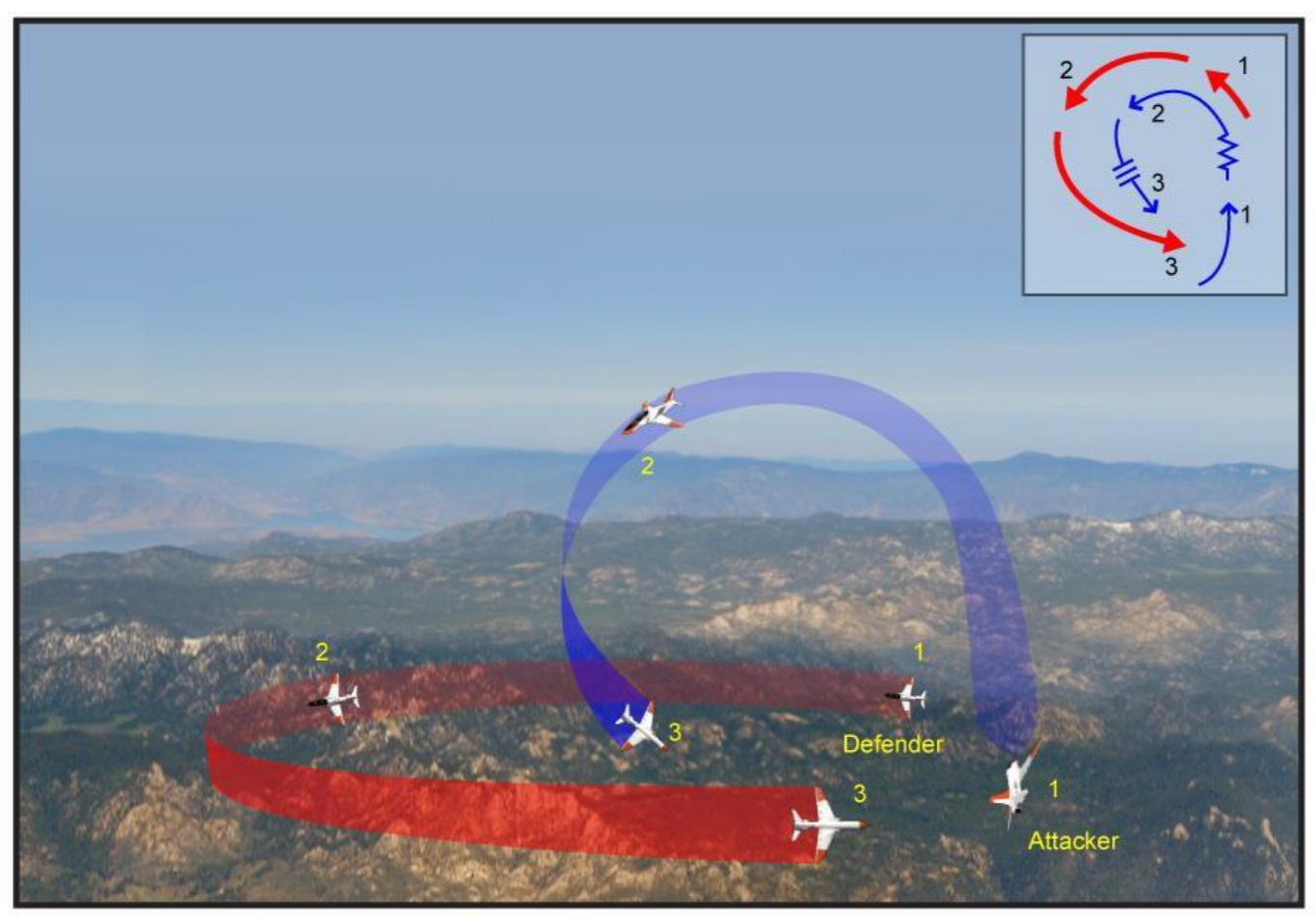}
\caption{Example of a high yo-yo maneuver from public domain CNATRA \cite{cnatraP826} training manual.}
\label{high_yoyo}
\vspace{-10pt}
\end{figure}

In this paper, we propose a pursuit/evasion problem formulation based on Markov Decision Processes (MDPs) and use our recently proposed algorithm \cite{bertram2019} to efficiently solve the problem even for large teams.  The algorithm seamlessly switches between pursuit and evasion while simultaneously avoiding collisions with other aircraft in the same team and the ground.  The algorithm is adaptable to multiple aircraft type through the use of forward projection of the aircraft dynamics, and a pseudo-6dof model is presented.

Our main contributions for this work are:
\begin{itemize}
    \item Extension of the 2D algorithm with discrete state space in \cite{bertram2019} to a continuous 3D state space;
    \item Addition of a forward projection module that allows the algorithm to support any arbitrary aircraft type;
    \item Demonstration of efficient algorithm performance that scales to large team. sizes
\end{itemize}

We additionally develop a 3D visualization tool to evaluate the algorithm and to provide insight to reviewers and readers on the complexity of the problem.

In Section II we identify and discuss related work.  In Section III we briefly provide background on Markov Decision Processes.  In Section IV we describe the method we use, including the pseudo-6DOF model and aircraft dynamics, as well as the details of the Markov Decision Process problem formulation used in this approach.  Section V describes the experimental setup to evaluate the pursuit/evasion contests, including the definition of two metrics we use to evaluate the behavior.  Section VI describes the results of our experiments demonstrating the efficient performance and consistent behavior of the algorithm.  Section VI also provides links to videos showing examples of varying sized teams competing against each other.

\section{Related Work}

There is extensive work from many communities which address different approaches to pursuit/evasion.  We describe several approaches and discuss how they relate to Markov Decision Process approach used in this paper.

Eklund etc. \cite{eklund2005implementing} described a nonlinear model predictive control (NMPC) approach to a pursuit/evasion problem using a set of cost functions with repulsive and attractive natures to shape the behavior of the pursuer.  An iterative optimization method was used to produce a solution at each time step using simplified aircraft dynamics.  Multiple matrices in the NMPC formulation required tuning to obtain good behavior.  It is worth noting that the cost functions used in their work are analogous to reward functions used for Markov Decision Processes.

Schopferer and Pfeifer \cite{schopferer2015performance} proposed a method to perform flight planning in the presence of a uniform wind field, with the aircraft motion modeled with trochoids.  The three dimensional flight path is constructed by superimposing a horizontal and vertical solution to obtain an approximate 3D path.  A probabilistic roadmap planner is used to generate global plans.

Vector fields approaches have also been used for pursuit/evasion problems.  Goncalves etc. \cite{gonccalves2010vector} described a vector field approach for convergence, circulation, and correction around a closed loop pattern.
Lawrence etc. \cite{lawrence2008lyapunov} presented a vector field approach for circular (or warped circular) patterns, and also describes a switching mechanism to handle waypoint following or arbitrary paths.  Stable tracking of the vector field is explored using Lyapunov techniques.  
Vector fields can be viewed as similar in nature to the optimal policy that is generated by solving a Markov Decision Process.  Where vector fields are generally applied over a continuous state space, MDP optimal policies normally describe actions that are intended to cause a transition from the current discrete state to a desired next discrete state.  

Within the robotics and computational geometry community, pursuit/evasion is often considered in a different context.  The pursuer(s) are attempting to search through an environment to observe the evader(s), similar to security guards searching through a museum for a potential intruder.  Often in these problem formulations, the goal is identifying the minimum number of pursuers needed in order to guarantee that if an evader is present within the environment that it will be detected, and is not focused on tracking or chasing the evader as in the target problem of this paper.  However, these works are instructive as the algorithm used in this paper is built on the recognition that an MDP can be represented as a graph.  Examples of this type of pursuit/evasion problem are \cite{guibas1997visibility, lavalle1997finding, kehagias2009graph}.  An example of graph based pursuit/evasion problem applied to graphs of infinite nodes is \cite{lehner2016pursuit}, where they describe the problem as a cop-and-robbers problem and define a winning strategy as preventing the robber from visiting a node in the infinite graph infinitely many times.  This allows strategies which either catch the robber or force the robber to flee `to infinity'.  
Markov Decision Processes are normally viewed as a tree of sequential actions, but can also be understood as a graph.  As most MDP problems normally have a discrete state space, this graph would normally also have a finite number of nodes.  Our method provides a way to support MDP problem formulations with continuous state spaces, and the corresponding graph would then have an infinite number of nodes.  Like the cop-and-robbers problem above, forcing an adversary to flee would be an acceptable strategy for our aircraft pursuit/evasion problem as well. 

Jia etc. \cite{shengde2014continuous} proposed a continuous-time Markov Decision Process (CTMDP) approach where variable time steps are allowed to be taken within a discretized state space where the transition function is defined instead as a transition rate function, allowing the possible resulting state transitions to be predicted with varied time steps.  The large state space is simplified by classifying the states into neutral, advantaged, disadvantaged, and mutually disadvantaged categories and a Bayesian method is used to determine the transition probabilities.  Pursuit/evasion within a 2D grid world environment is considered.  

Within the optimal control community, one area of related work is Differential Dynamic Programming (DDP) which uses dynamic programming to iteratively improve a local optimal control policy.  Sun etc. \cite{sun2018min} used DDP to solve an adversarial aircraft pursuit/evasion problem, terming their approach as game-theoretic DDP (GT-DDP) by combining DDP with a min-max problem formulation.  Differential Dynamic Programming and Markov Decision Processes have much in common and both stem from Bellman's original work on dynamic programming \cite{bellman1957dynamic}.  Where the optimal control field focuses on the Hamilton-Jacobi-Bellman (HJB) equation and differentiable dynamics, MDPs often generalize the dynamics into a (deterministic or stochastic) transition function which captures uncertainty about the environment through probabilities (similar to those used for Markov chains.)  Comparing \cite{sun2018min} to this paper's work, GT-DDP in \cite{sun2018min} does have a much richer capability to incorporate system dynamics, but this comes at the expense of additional computation time and a need for convergence of the iterative nature of the algorithm.  

The most relevant paper to this work is \cite{mcgrew2010air} which describes a Markov Decision Process based pursuit/evasion problem for aircraft using approximate dynamic programming.  A state space was formed from a set of features which minimized mean squared error using a forward-backward search.  Trajectory sampling was used to obtain training data that would be likely to have value during training.  Reward shaping was used to guide the exploration to the desired behavior in the form of a scoring function heuristic developed by an expert.  Rollout was used to extract a refined policy from the approximation computed via approximate dynamic programming (ADP) and was accelerated with a neural net.  The dynamics model for the airplane used is a Dubin's airplane without any vertical components or altitude modeled.  

There are some subtle differences between this paper and the work in \cite{mcgrew2010air}.  \cite{mcgrew2010air} is a good example of using a variety of practical techniques to deal with the intractability of large MDP state spaces, whereas this work explicitly uses a state space designed to be intractable by traditional MDP methods via the use of a continuous state space resulting in an MDP with an infinite number of states in order to demonstrate scaling to continuous state spaces.  \cite{mcgrew2010air} uses a 2D aircraft model, where this paper uses a 3D pseudo-6DOf model to demonstrate scaling to a continuous 3D state space and to demonstrate full maneuvering by the aircraft (e.g., loops, rolls, spirals).  In this paper, no reward shaping is required to speed up or aid convergence, as the underlying MDP is solved directly without relying on typical methods used for approximate dynamic programming. And finally, in \cite{mcgrew2010air} 1v1 pursuit/evasion is explored where in this paper scaling to 10v10 teams is demonstrated.

Also of note are \cite{park2016differential} and \cite{zhang2018research}.  Park etc. \cite{park2016differential} used a higher fidelity 3D model and a min-max approach over a sliding window to demonstrate 1 vs 1 pursuit/evasion, and while the behavior in simulation appears promising, the real-time performance of the algorithm is not reported.  In \cite{zhang2018research}, a reinforcement learning approach is taken using deep Q-learning using a 2-layer multi-layer perceptron as the function approximator, and with a modified epsilon-greedy exploration strategy where a heuristic function used in place of random action in order to avoid wasteful actions during exploration.  Performance is examined in 2D.

\section{Background}

Markov Decision Processes (MDPs) are a framework for sequential decision making with broad applications to finance, robotics, operations research and many other domains \cite{suttonbarto}.  MDPs are formulated as the tuple $(s_t, a_t, r_t, t)$ where $s_t \in S$ is the state at a given time $t$, $a_t \in A$ is the action taken by the agent at time $t$ as a result of the decision process, $r_t$ is the reward received by the agent as a result of taking the action $a_t$ from $s_t$ and arriving at $s_{t+1}$, and $T(s_t, a, s_{t+1})$ is a transition function that describes the dynamics of the environment and capture the probability $p( s_{t+1} | s_t, a_t )$ of transitioning to a state $s_{t+1}$ given the action $a_t$ taken from state $s_t$.  
    
A policy $\pi$ can be defined that maps each state $s \in S$ to an action $a \in A$.  From a given policy $\pi \in \Pi$ a value function $V^\pi(s)$ can be computed that computes the expected return that will be obtained within the environment by following the policy $\pi$.  The value function can be expressed in the iterative Bellman equation as follows, where $r_{t+1}$ represents immediate reward collected by taking an action $a_t$ which leads to a next state $s_{t+1}$ and a value of $V( s_{t+1})$.  This the value function for any state is the current reward plus the discounted future reward that can be obtained by taking the best action from the current state, and is an expectation of the future reward that can be obtained from the current state.

\begin{align}
    V(s) = r_{t+1} + \gamma \max_{a_t} V( s_{t+1} )
\end{align}
    
The solution of an MDP is termed the optimal policy $\pi^*$, which defines the optimal action $a^* \in A$ that can be taken from each state $s \in S$ to maximize the expected return.  From this optimal policy $\pi^*$ the optimal value function $V^*(s)$ can be computed which describes the maximum expected value that can be obtained from each state $s \in S$.  And from the optimal value function $V^*(s)$, the optimal policy $\pi^*$ can also easily be extracted.

\section{Method}

We use the algorithm described in \cite{bertram2019} as the underlying guidance and collision avoidance algorithm which demonstrated collision avoidance in a 2D environment.  The algorithm is extremely efficient and the paper demonstrated good performance on a discretized state space.  We extend the method to demonstrate performance in a continuous state space while also extending it to a 3D environment to demonstrate scaling to the higher dimensional space.  Demonstration of scaling is further highlighted by showing large teams performing pursuit/evasion together.  Finally, we introduce a pseudo-6DOF model allowing the aircraft to roll, pitch, and perform complex aerial maneuvers which serves to further demonstrate the power of this approach.

\subsection{Dynamic Model}

The aircraft kinematic model is a pseudo 6 degree of freedom (pseudo-6DOF) model which approximates fixed wing aircraft motion given inputs similar to stick and throttle inputs.  The model provides a way to study the algorithms behavior without requiring full aerodynamics to be modelled.  The algorithm needs this pseudo-6DOF model to provide ``forward prediction".  This means that from a given current state, the model must be able to calculate the future state of applying a given set of possible control actions for a fixed number of timesteps.  Any model which satisfies this requirement can be integrated with the algorithm, including full-fidelity 6DOF fixed-wing models, helicopters, quad rotors, and models with underlying autopilot controllers.

The model used is an extension of the pseudo-6DOF formulation in \cite{park2016differential} and also incorporates a few additional terms in the model in \cite{huynh1987numerical}. It should be considered as a simplified model of \cite{huynh1987numerical}.

\begin{itemize}
    \item $n_x$: Throttle acceleration directed out the nose of the aircraft in $g$'s
    \item $V$: Airspeed in meters/second.
    \item $\gamma$: Flight path angle in radians.
    \item $x, y, z$: position in NED coordinates in meters where altitude $h=-z$
    \item $\phi$: Roll angle in radians
    \item $\psi$: Horizontal azimuth angle  in radians
    \item $\alpha$: Angle of attack in radians with respect to the flight path vector
\end{itemize}

The inputs to the model are: (1) the thrust $n_x$, (2) the rate of change of angle of attack $\dot{\alpha}$ and (3) the rate of change of the roll angle $\dot{\phi}$.

The equations of motion for the aircraft are:
\begin{align}
   \dot{V} &= g \left[ n_x \cos \alpha - \sin \gamma \right], \\
   \dot{\gamma} &= \frac{g}{V} \left[ n_f \cos \phi - \cos \gamma \right], \\
   \dot{\psi} &= g \left[ \frac{ n_f \sin \phi }{ V \cos \gamma } \right],
\end{align}
where the acceleration exerted out the top of the aircraft $n_f$ in $g$s is defined as:
\begin{align}
    n_f &= n_x \sin \alpha + L ,
\end{align}
with a lift acceleration of $L=0.5$.  Here, 1 ``g" is a unit of acceleration equivalent to $9.8 ~m/{s^2}$.   $L$ was chosen to provide some amount of lift while in flight to partially counteract gravity and provide a stable flight condition with a low positive $\alpha$ angle of attack in the pseudo-6dof model.  For a true aerodynamic model, this lift varies by the velocity (Mach number), but this level of detail is omitted in our simplified pseudo-6dof.

The kinematic equations are:
\begin{align}
   \dot{x} &= V \cos \gamma \cos \psi \\
   \dot{y} &= V \cos \gamma \sin \psi \\
   \dot{z} &= V \sin \gamma. 
\end{align}

While this model is not aerodynamically comprehensive, it is sufficient to describe aircraft motion suitable for examining the algorithm behavior without loss of generality. Again, our algorithm can integrate with any aircraft dynamic model that provides a forward prediction.

\subsubsection{Forward Projection} \label{forward_projection}

In order to determine the future state resulted from a given action, we use forward projection to simulate the dynamics forward in time.  We use a discrete time step of $0.1$ seconds and apply the control actions at each time step for a specified number of time steps.

For the purposes of determining the future state of an action, we forward project for 1 time step (0.1 second).  After selecting an action and applying it to the simulation, we advance the simulation one time step (0.1 seconds).  Thus an action is chosen at a 10 Hz rate with a 1 second forward projection horizon. 

The simulated future states can be viewed as an approximation of the reachable states, and are applied to the solution of the Markov Decision Process (MDP) to determine the value of the potential future states the agent might reach.  Thus the agent follows the optimal policy of the MDP at each time step by determining which future reachable state is most valuable, and then takes the action in the next time step that will lead it towards that state.

Each team is provided with different aircraft performance limits which serve to provide the ``blue" team (team 0) with a performance advantage over the ``red" team (team 1) and prevents deadlocks where neither team is able to obtain an advantage over the other.  
Table \ref{team_limits} lists the performance limits, where the speed of sound $Mach = 343 ~m/s$.  These limits were chosen to represent a highly maneuverable subsonic UAV and do not represent any real aircraft.

\begin{table}[h]
\caption{Limits on aircraft performance for each team}
\label{team_limits}
\begin{center}
\begin{tabular}{ |c|c|c|c|c|c|c| } 
 \hline
 & & & & & & \\
 Team & $V_{min}$ & $V_{max}$ & $\dot{\psi}_{\text{min}}$ & $\dot{\psi}_{\text{max}}$ & $\alpha_{\text{min}}$ & $\alpha_{\text{max}}$\\
  & (Mach)  & (Mach) & (rad/s) & (rad/s) & (rad) & (rad) \\
 & & & & & & \\
 \hline
 Blue  & 0.1 & 0.35 & -1.5 & -1.5 & -.009 & .69 \\
 Red &  0.1 & 0.30 & -1.3 & -1.3 & -.009 & .52  \\
 \hline
\end{tabular}
\end{center}
\end{table}

\subsection{MDP Formulation}

\subsubsection{State Space}

We define the environment where the aircraft operates within a 25 km by 25 km by 25 km volume which is treated as a continuous state space.  There are two teams of aircraft in this environment: a ``blue" team and a ``red" team.  Each aircraft (an ``ownship'') is controlled by our proposed algorithm, and aircraft on the blue team have a slight performance advantage over aircraft on the red team.

The state includes all the information each ownship needs for its
decision making: the full aircraft state of the ownship, the position and velocity of every teammate aircraft, and the position and velocity of every opponent aircraft.  

Each ownship is aware of its own aircraft state produced by the pseudo-6DOF model.  For each ownship, the state is formed by concatenating the following:  
\begin{itemize}
    \item  $\zeta$ the pseudo-6DOF state: position $x, y, z$, the heading angle $\psi$, the roll angle $\phi$, the flight path angle $\gamma$, the pitch angle $\theta$, the angle of attack $\alpha$, and the speed $V$.   
    \item for each teammate $f_j,  \forall j \in J$: the position $f_{j,x}, f_{j,y}, f_{j,z}$ and velocity $f_{j,v_x}, f_{j,v_y}, f_{j,v_z}$, and 
    \item for each opponent aircraft $i_k, \forall k \in K$: the position $i_{k,x}, i_{k,y}, i_{k,z}$ and velocity $i_{k,v_x}, i_{k,v_y}, i_{k,v_z}$
\end{itemize}
\begin{align}
s_o = [ \zeta,  
f_1, \cdots, f_j, 
i_1, \cdots, i_m ]
\end{align}
where $j$ represents the number of teammates, and $m$ represents the number of opponents.

\subsubsection{Action Space}

Inputs to the model are (1) the thrust $n_x$, (2) the rate of change of angle of attack $\dot{\alpha}$ and (3) the rate of change of the roll angle $\dot{\phi}$.

The action space is then:
\begin{align}
A = \{ \dot{\alpha}, \dot{\phi}, n_x \}.
\end{align}

There are two teams of aircraft $k \in \{ 0, 1 \}$ where team $k=0$ is the ``blue team" and $k=1$ is the ``red team".  When the teams' aircraft have equivalent performance, simulations often result in a stalemate which represent a Nash equilibrium where neither aircraft is able to gain advantage over the other.  In these cases, simulation will not naturally terminate.  Therefore, in the simulations we provide a performance advantage to the blue team which more naturally leads to simulations that terminate.

\begin{table}[h]
\caption{Action choices for each team}
\label{team_limits}
\begin{center}
\begin{tabular}{ |c|c|c|c| } 
 \hline
 & & & \\
 Team & $\dot{\phi}$ & $\dot{\alpha}$ & $n_x$ \\
  & (rad/s)  & (rad/s) & (g's) \\
 %& & &  \\
 \hline
 Red  & -1, -.8, $\cdots$, .8, 1 & -.5, -.4, $\cdots$, .4, .5  &  0, 1, $\cdots$, 6  \\
 Blue  & -1.5, -1.2, $\cdots$, 1.2, 1.5 &  -.5, -.4, $\cdots$, .4, .5  &  0, 1, $\cdots$, 8  \\
 \hline
\end{tabular}
\end{center}
\end{table}

\subsubsection{Reward Function} \label{reward_function}

The primary mechanism to control the behavior of an agent in a Markov Decision Process (MDP) is through the Reward Function.  By providing positive and negative rewards to the agent, it is able to determine which actions lead to positive reward and the solution of an MDP maximizes the expectation of future reward. 
In our pursuit evasion problem, we will use positive and negative rewards that are coupled together to create tension between potential actions.  For example, we will place a positive reward near the location of an aircraft to attract other aircraft, but we will also place a negative reward at the aircraft to prevent a collision.  A natural equilibrium develops between these positive and negative rewards that generates the desired behavior of approaching another aircraft without colliding with it.

Following the approach used by Bertram et al. in \cite{bertram2019}, we will treat each negative reward as a ``risk well", which is a region of negative reward (i.e., a penalty) which is more intense at the center and decays outward until a fixed radius is reached, where after no penalty is applied.  
We present our reward function in terms of the behaviors we wish to obtain in Table \ref{reward_table}.  In this table, $\hat{p}$ represents the current position of an aircraft (teammate or opponent) and $\hat{v}$ represents that aircraft's current linear velocity.  In some cases we project the aircraft's position forward in time with an expression $\hat{p} + \hat{v} t$ and then define a range of time as in $\forall t \in \{0, 1, 2\}$ to indicate that we create a reward at the location of the aircraft at each timestep in the future indicated by the range of $t$.

\begin{table*}[t]
\vspace{5pt}
\caption{Rewards created for each ownship}
\label{reward_table}
\begin{center}
\begin{tabular}{ |c|c|c|c|c|l| } 
 \multicolumn{6}{l}{\textit{For each teammate:}} 
 \vspace{4pt}
 \\
 %\multicolumn{6}{l}{} \\
 \hline
 \rowcolor{Gray}
 Magnitude  & Decay factor & Location & Radius & Timesteps & Comment \\
 \hline
 $-100$ & $.97$ & $\hat{p} + \hat{v} t $ & $150 + 10 t$ & $\forall t \in \{ 0, 1, 2, 3, 4, 5\}$  & Collision avoidance, 5 rewards \\
 $10$ & $.999$ & $\hat{p}$ & $\infty$ & N/A & Weak formation flight or clustering\\
 \hline
 
 \multicolumn{6}{l}{} \\
 \multicolumn{6}{l}{\textit{For each opponent:}} 
 \vspace{4pt}
 \\
 %\multicolumn{6}{l}{} \\
 \hline
 \rowcolor{Gray}
 Magnitude  & Decay factor & Location & Radius & Timesteps & Comment \\
 \hline
 $-300$ & $.99$ & $\hat{p} + \hat{v} t $ & $\hat{v} t$ & $\forall t \in \{ 0, 1, 5, 10\}$  & Collision avoidance, 4 rewards \\
 $200$ & $.999$ & $\hat{p}$ & $\infty$ & N/A  & Pursuit \\
 \hline
\end{tabular}
\label{team_size_results}
\end{center}
\vspace{-15pt}
\end{table*}

All aircraft also receive a penalty below a certain altitude which prevents the aircraft from plummeting into the terrain.  For this paper, $h_{\text{max}}$ is the maximum height of the terrain that is loaded into the simulation.  We define a minimum safe altitude known as the ``hard deck" in which we will allow the aircraft to fly.  Any aircraft which goes below the hard deck for the purposes of the game has crashed and is removed from the simulation.  We define the hard deck $h_{\text{deck}} = h_{\text{max}} + 500$.
For any state with an altitude of $h$ from the hard deck up to an altitude of $h_{\text{penalty}} = h_{\text{deck}} + 1000$, a penalty is applied $r_{\text{penalty}} = -( 10000 - h)$ which is a very strong negative reward that will override any other positive rewards in the game.

\subsection{Algorithm}

We alter the algorithm from \cite{bertram2019} by extending it to handle 3D aircraft positions in a continuous state space and by adapting it to allow for forward projection.  We present the algorithm which we call \fastmdp in Algorithm \ref{fastmdp_algo}.  Note that the algorithm is presented for clarity here; when implemented certain optimizations are made which improve performance but make it more difficult to understand.

\newcommand{\CommentLine}[1]{
    \vspace{2pt}
    \State // \textit{#1}
    \vspace{2pt}
}

\newcommand{\norm}[1]{\left\lVert#1\right\rVert}

%%%%%%%%%%%%%%%%%%%%%%%%%%%%%%%%%%%%%%%%%%%%%%%%%%%%%%%%%%
%\noindent
%\begin{minipage}[t]{\columnwidth}
%    \vspace{-5pt}
\begin{algorithm}[tbph]
    \footnotesize
    \caption{Pursuit Evasion with \fastmdp} \label{fastmdp_algo}
    \begin{algorithmic}[1]
    \Procedure{Pursuit Evasion}{$\textit{ownshipState}, ~ \textit{worldState}$}
        \State $worldState \gets$ randomized initial aircraft states
        \State $actions \gets$ \textit{list of actions for ownship's team (precomputed)}
        \State $nextWorldState \gets$ allocated space 
        \While {both teams have aircraft remaining}
            \For{each ownship}
                \State $currState \gets worldState[ownship]$
                \vspace{2pt}
                \State // {Build peaks per Table \ref{reward_table}}
                \vspace{2pt}
                \State $\textit{posPeaks} \gets $\textit{build pos rewards}
                \State $\textit{negPeaks} \gets $\textit{build neg rewards in Standard Positive Form}
                \vspace{2pt}
                \State // Perform forward projection per Section \ref{forward_projection}
                \vspace{2pt}
                \State $\textit{oneStep} \gets fwdProject(currState, actions, 0.1 ~s)$
                \State $\textit{reachStates} \gets fwdProject(currState, actions, 1.0 ~s)$
                \vspace{2pt}
                \State // Compute the value at each reachable state
                \vspace{2pt}
                \State $trueVals \gets $ space for each state
                \For{$state \in reachStates$}
                    \vspace{2pt}
                    \State // First for positive peaks 
                    \vspace{2pt}
                    \For{$p_i = posPeak_i \in posPeaks$}
                        \State $d_p \gets \norm{ state - \mathbf{location}(p_i) } _2 $ 
                        \Comment{distance}
                        \State $r_p \gets \mathbf{reward}(p_i)$
                        \State $\gamma_p \gets \mathbf{discount}(p_i)$
                        \State $posValues_i \gets |r_p|| \cdot \gamma_p ^ {d_p} $
                    \EndFor
                    \State $posMax \gets \underset{i}{\max} ~ posValues_i $
                    \vspace{2pt}
                    \State // Next for negative peaks (in Standard Positive Form)
                    \vspace{2pt}
                    \For{$n_i = negPeak_i \in negPeaks$}
                        \State $d_n \gets \norm{ state - \mathbf{location}(n_i) } _2 $ 
                        \Comment{distance}
                        \State $\rho_n \gets negDist_i < \mathbf{radius}(n_i) $ \Comment{within radius}
                        \State $r_n \gets \mathbf{reward}(n_i)$
                        \State $\gamma_n \gets \mathbf{discount}(n_i)$
                        \State $negValues_i \gets int(\rho_n) \cdot |r_n|| \cdot \gamma_n ^ {d_n} $
                    \EndFor
                    \State $negMax \gets \underset{i}{\max} ~negValues $
                    \vspace{2pt}
                    \State // Hard deck penalty
                    \vspace{2pt}
                    \If{$\mathbf{altitude}(state) < penaltyAlt$}
                        \State $hDeck \gets 1000 - \mathbf{altitude}(state)$
                    \Else
                        \State $hDeck \gets 0$
                    \EndIf
                    \State $trueVals[state] \gets posMax - negMax - hDeck$
                \EndFor
                
                \vspace{2pt}
                \State // Identify the most valuable action
                \vspace{2pt}
                \State $bestActionIdx \gets \argmax(trueVals)$
                \vspace{2pt}
                \State // For illustration, the corresponding value
                \vspace{2pt}
                \State $maxValue \gets trueVals[ bestActionIdx ]$
                \vspace{2pt}
                \State // And the next state when taking the action
                \vspace{2pt}
                \State $nextState \gets oneStep[ bestActionIdx ]$
                
                \State $nextWorldState[ ownship ] \gets nextState$ 
            \EndFor
            
            \State // Now that all aircraft have selected an action, apply it
            \vspace{2pt}
            \State $worldState \gets nextWorldState$
        \EndWhile
    \EndProcedure
    \end{algorithmic}
    \label{fastmdp}
    \end{algorithm}

In order to efficiently solve the MDP, the algorithm from \cite{bertram2019} divides the rewards into positive and negative rewards and processes them separately.  The positive rewards are processed in a straightforward manner where each reward is treated as a peak which decays exponentially, and the resulting value function surface is the $\max$ of all of these exponentially decaying peaks.  Negative rewards are treated differently.   They are first converted to what Bertram et al. refers to as ``Standard Positive Form" (S.P.F.) where each negative reward is negated so that each negative reward becomes a positive reward in this standard positive form space.  Once in standard positive form, a new value function surface is computed from the rewards in the same manner as the positive rewards.  The value function surface in standard positive form is then negated, resulting in a value function surface that is negative and is a close approximation of solving the MDP with only the negative rewards present.  When this surface resulting from the negative rewards is summed with the surface resulting from the positive rewards, the result is the value function surface that closely approximates the result that would be obtained from solving the original MDPs with all rewards present. 
We reuse this basic approach, but employ some additional computational optimizations to make these operations more efficient. 

All of these steps are optimized as much as possible for operation on a CPU.  As the code is implemented in Python, an optimization library known as numba is employed which recompiles key sections of the code as C code to obtain faster operation.  Additionally, the code is written to take advantage of the numerical library numpy to perform vectorized operations over arrays.  No parallelization on CPU via multiple cores or GPU are employed.

\phantom{line}

\section{Experimental Setup}

We demonstrate this MDP based planner in a 3D aircraft simulation showing a view of the two teams of aircraft.
The simulation covers a configurable sized volume which contains a configurable number of team members on each of the two teams.  

Simulation begins with both teams spawned randomly on opposing sides of the environment.  The teams must each avoid collisions with team mates while simultaneously pursuing members of the opposing team using only the reward system we have defined above.

At each time step, the simulation generates the state updates for each ownship.  Each ownship creates and solves its own MDP using the highly efficient algorithm presented in \cite{bertram2019}.  Each ownship forward projects each possible action by 1 second, and then uses the solution of the MDP to determine which action results in the highest valued future state.  The action selected with this method will then be applied in simulation for 1 timestamp (0.1 seconds).  The actions of all aircraft from both sides are selected and performed simultaneously without knowing the selected actions of any other aircraft in the simulation.   Simulation then advances by one time step.  Note that a new MDP is calculated at each time step, which is made possible by the performance of the algorithm in \cite{bertram2019}.

In this pursuit/evasion game, we define a pursuer ``capturing" an opponent if it is in a certain region behind the evading aircraft.  The ``control point" is defined as the position the evader was at 3 seconds previously.  If the pursuer is within 100 meters of the control point and relative angle between the two velocity vectors of the aircraft is within 60 degrees, then the pursuer is close to the control point and pointing at the evader and we consider this a sufficient condition for the pursuer to be able to ``capture" the evader (e.g., within range of some weapon).  The pursuer must maintain this condition for 30 consecutive time steps in order to successfully ``hit" the evader, which is analogous to a weapon taking some time to track the evader.  This is indicated visually in the simulation as a red pulsing rectangle around an aircraft that is in danger of being captured.

We build a scoring system that tracks the number of airplanes that have been captured.  When a team's airplane is captured, the opposing team is awarded one point.  Thus complete success is when one team reaches a score that equals the number of airplanes on the opposing team.  A ``win" is described as one team scoring higher than the other, with the other team necessarily incurring a ``loss", and a ``draw" is when both teams score the same.

We define a metric $P_{\text{win}}$ to study the effect of the algorithm over $N$ runs which is defined for a team as the number of wins the team obtained $W$ over the number of runs:  $P_{\text{win}} = \frac{ W }{ N }$. 
This metric can be applied to 1 vs 1 encounters and can scale to larger teams as well.

The $P_{\text{win}}$ measurement alone is not sufficient.  Beyond the probability of win, we also wish to define a metric that describes the survivability of the team.  In a 10 vs 10 game, it is clearly better when when winning if all 10 of the teammates survive as compared to a win when only 1 of the teammates remain at the end.  If we define the number of aircraft at the beginning of the contest as $N_{t_0}$ and the number remaining at the end of the contest as $N_{t_f}$, then we can define the ratio of teammates that survived a given contest $i$ as $P_{s_i}= {N_{t_f}}/{N_{t_0}}$.   Over $m$ contests, we define the overall probability of survivibility as $P_s = \frac{1}{m} \sum_{i=1}^{m} P_{s_i}$ where $m$ is the number of contests and is the average probability that the team will survive the contest.

\begin{figure}[tbp]
\vspace{5pt}
\centering
\begin{subfigure}{.95\columnwidth}
  \centering
  \includegraphics[width=\textwidth]{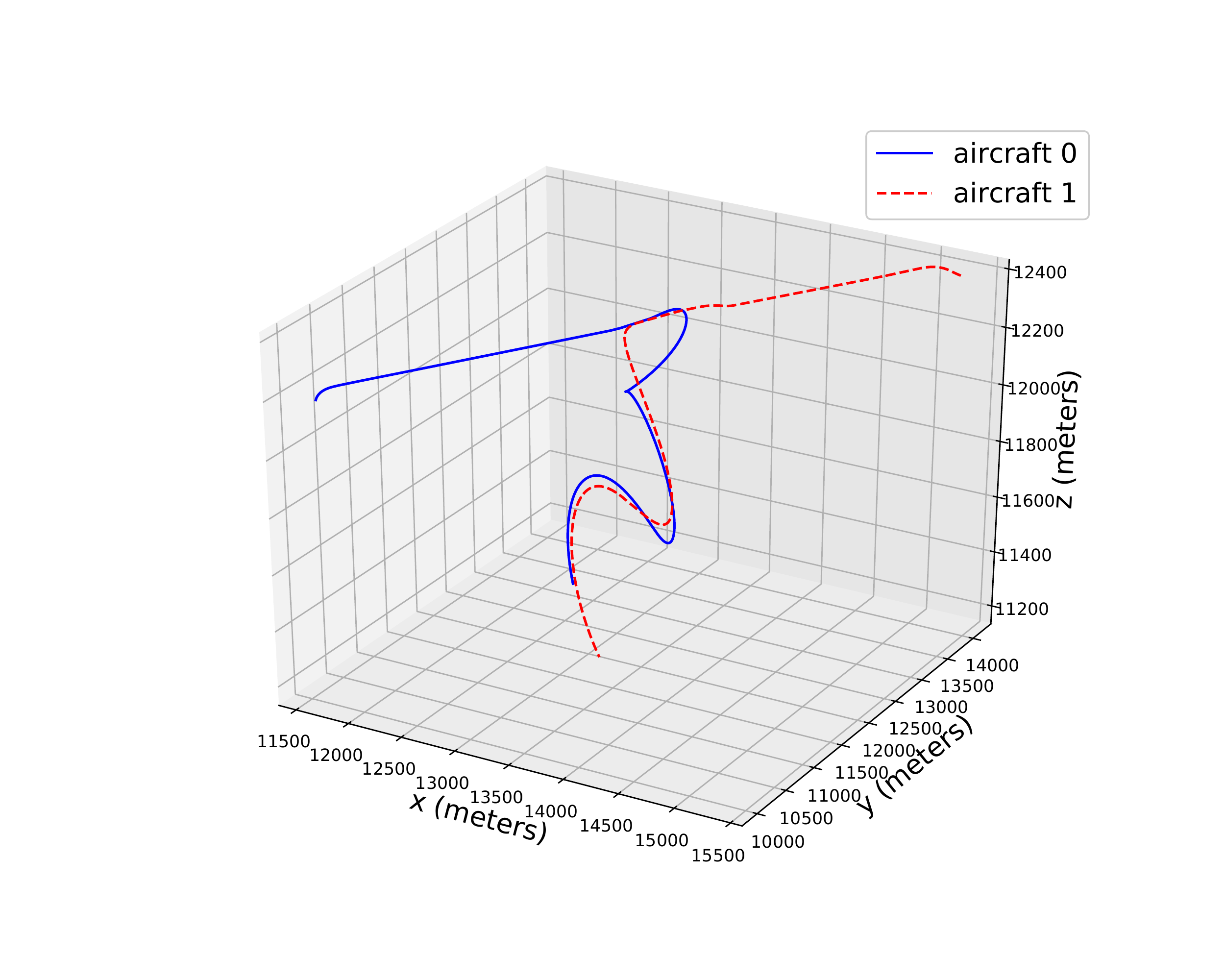}
  \caption{Trajectory of a sample 1v1 pursuit/evasion run}
\end{subfigure}
\begin{subfigure}{.95\columnwidth}
  \centering
  \includegraphics[width=\textwidth]{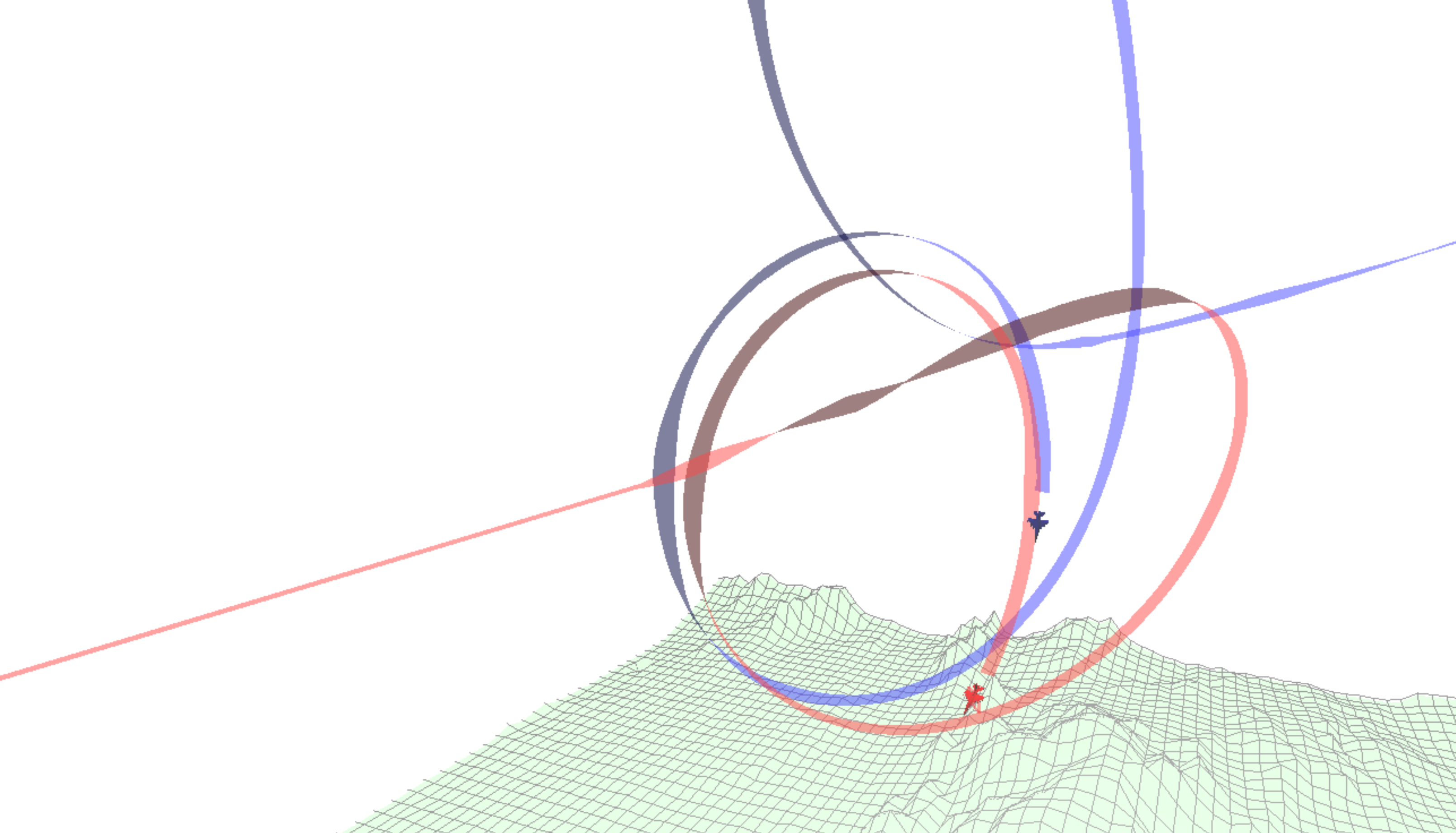}
  \caption{The same 1v1 run in a 3D visualization}
\end{subfigure}
\caption{Experimental results showing the performance of the algorithm for a 1v1 pursuit/evasion run.  (a) shows the trajectories of two aircraft in a standard Matlab style plot.  (b) shows the trajectories in a 3D visualization developed for this paper where ribbons are used to show historical attitude a 3D aircraft is used to more readily show current aircraft attitude.  Links to videos are provided for the interested reader in the results sections.}
\label{plot_1v1_trajectory}
\end{figure}

\begin{figure}[tbp]
\centering
\begin{subfigure}{.95\columnwidth}
  \centering
  \includegraphics[width=.95\columnwidth]{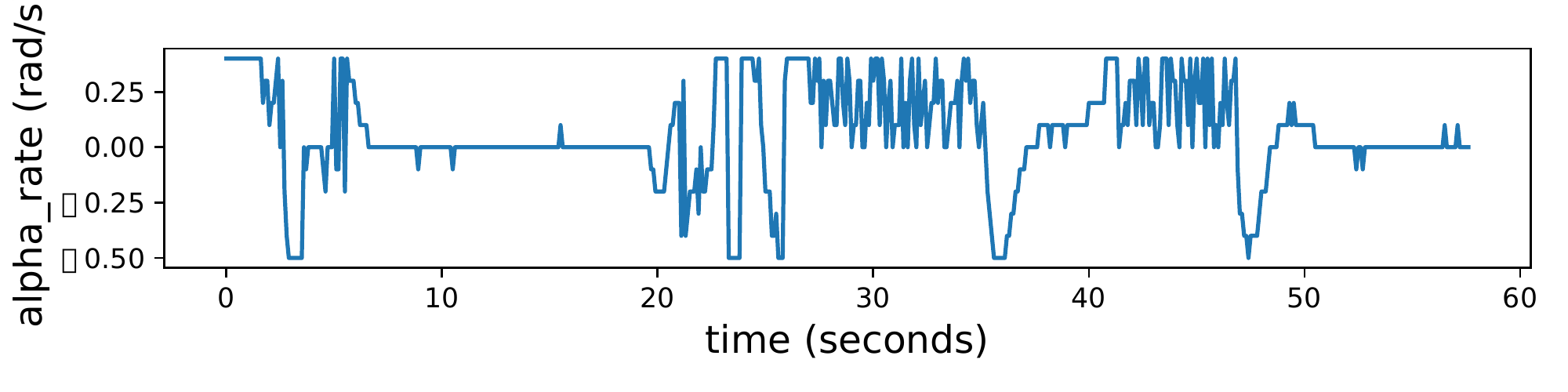}
\end{subfigure}
\begin{subfigure}{.95\columnwidth}
  \centering
  \includegraphics[width=\columnwidth]{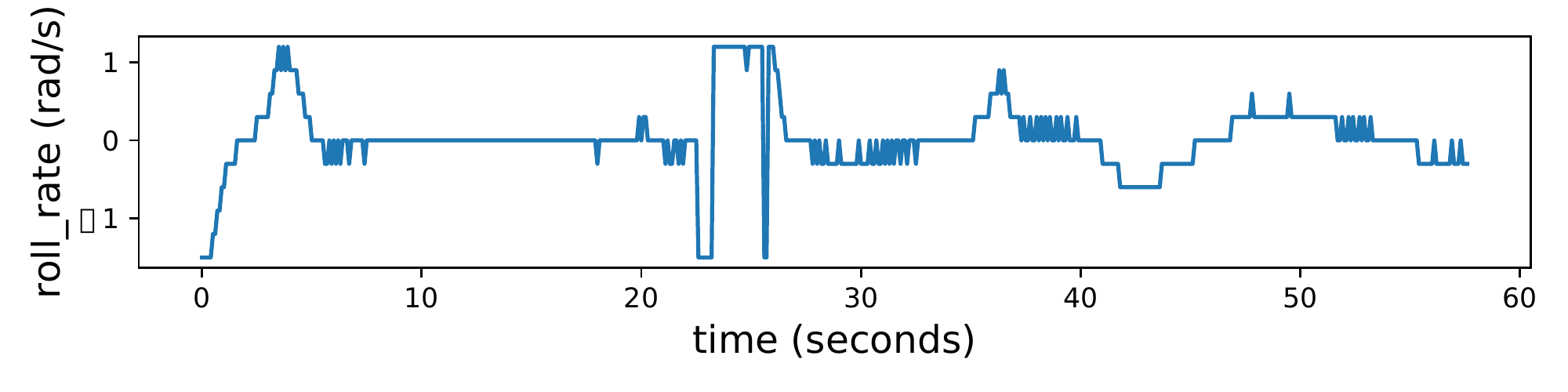}
\end{subfigure}
\begin{subfigure}{.95\columnwidth}
  \centering
  \includegraphics[width=.95\columnwidth]{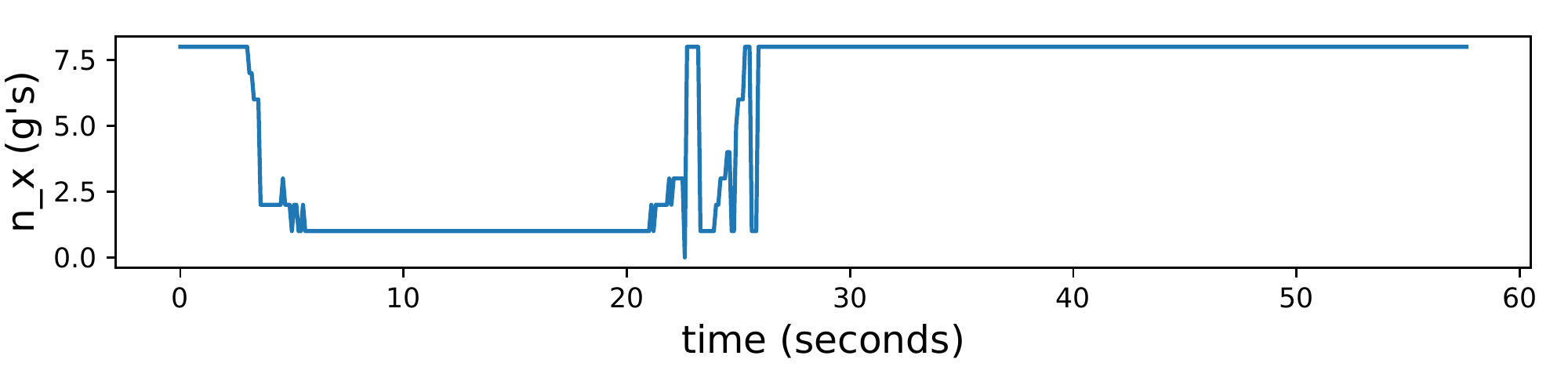}
\end{subfigure}
\caption{Experimental results showing the actions taken by the pursuer (blue aircraft) over time.  Alpha rate here is analogous to pushing forward or pulling back on the stick.  Roll rate is analogous to moving the stick from side to side.  $n_x$ is analogous to a throttle setting.}
\label{plot_1v1_actions}
\end{figure}

\begin{figure}[tbp]
\centering
\begin{subfigure}{.95\columnwidth}
  \centering
  \includegraphics[width=.95\columnwidth]{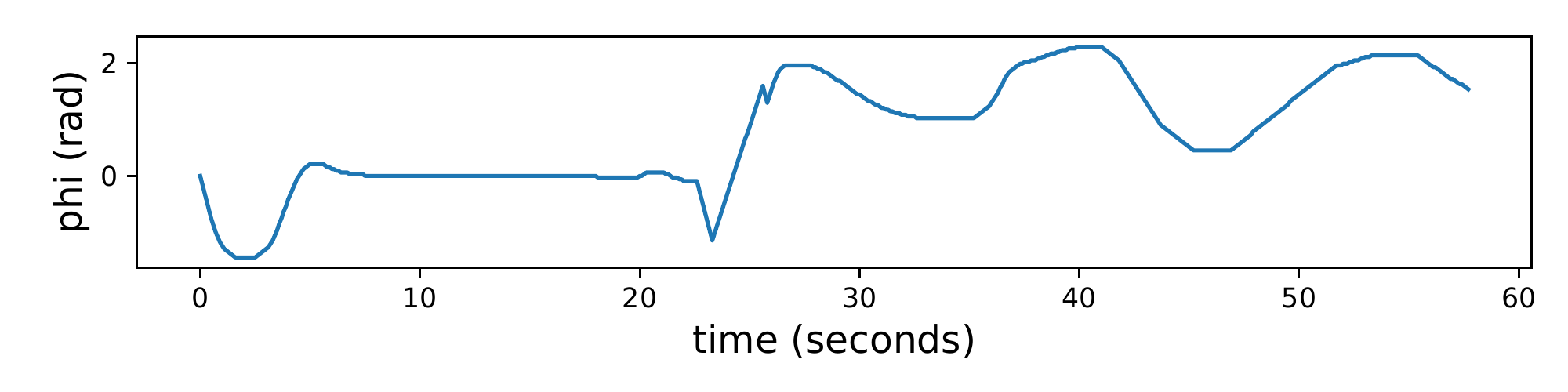}
\end{subfigure}
\begin{subfigure}{.95\columnwidth}
  \centering
  \includegraphics[width=.95\columnwidth]{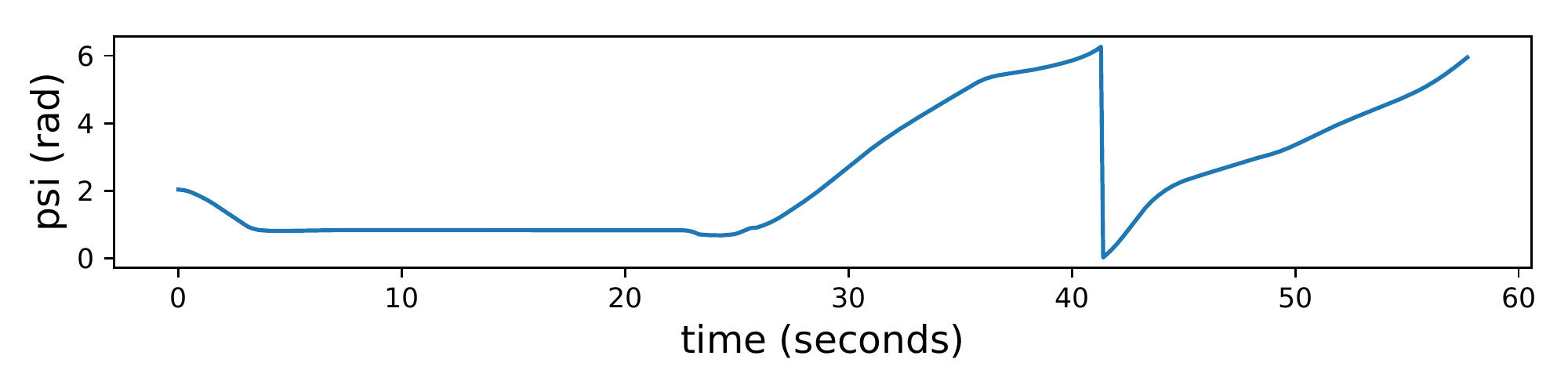}
\end{subfigure}
\begin{subfigure}{.95\columnwidth}
  \centering
  \includegraphics[width=.95\columnwidth]{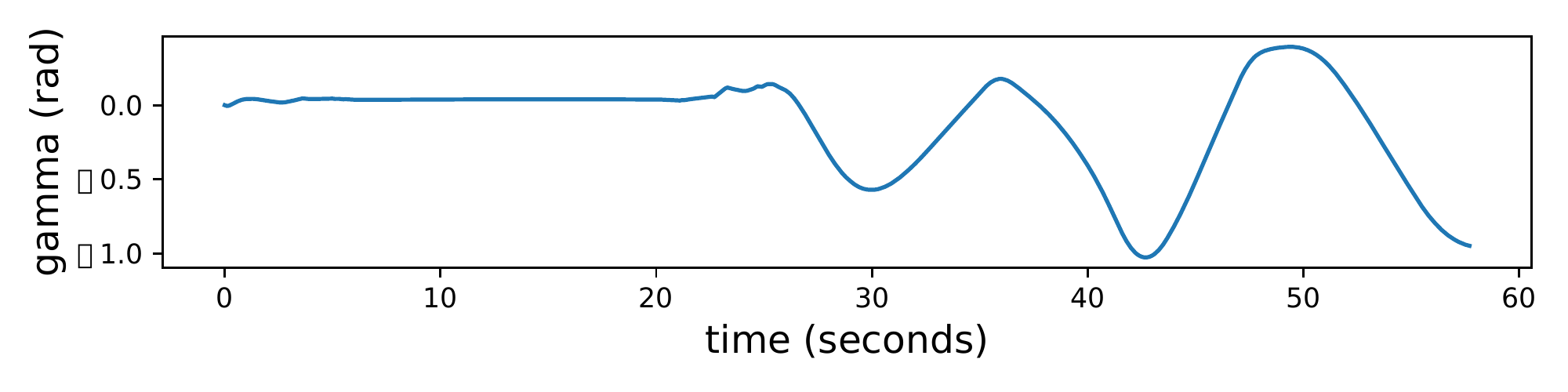}
\end{subfigure}
\begin{subfigure}{.95\columnwidth}
  \centering
  \includegraphics[width=.95\columnwidth]{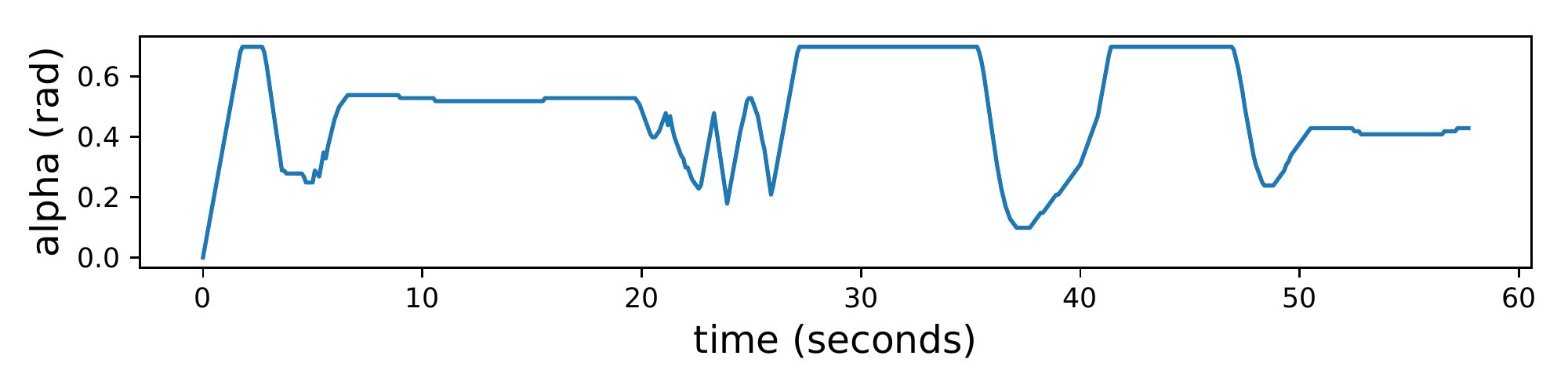}
\end{subfigure}
\begin{subfigure}{.95\columnwidth}
  \centering
  \includegraphics[width=.95\columnwidth]{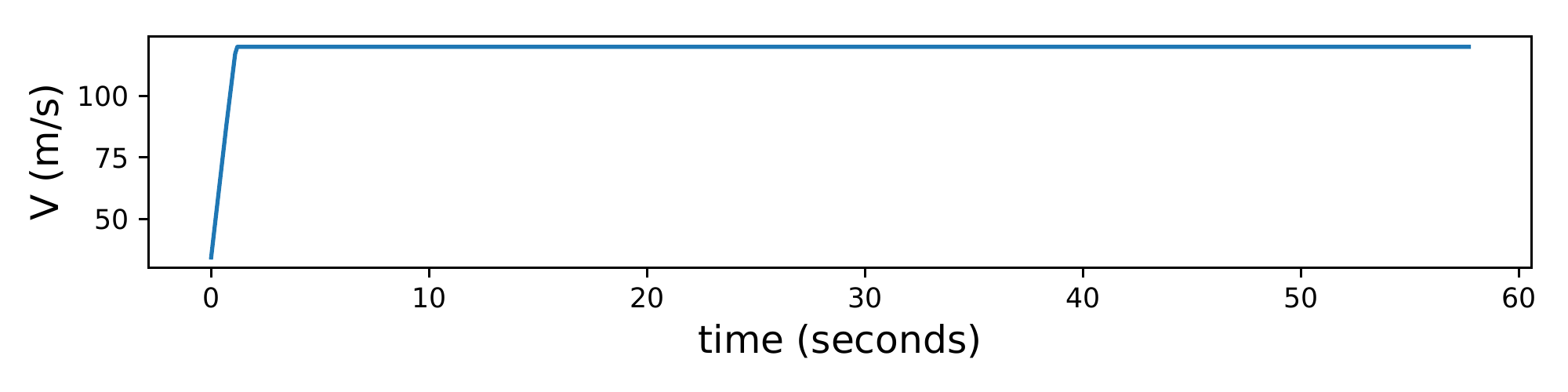}
\end{subfigure}
\caption{Experimental results showing the dynamics of the pursuer (blue aircraft) over time.}
\label{plot_1v1_dynamics}
\end{figure}

\begin{figure}[tbp]
\centering
\vspace{5pt}
  \includegraphics[width=.95\columnwidth]{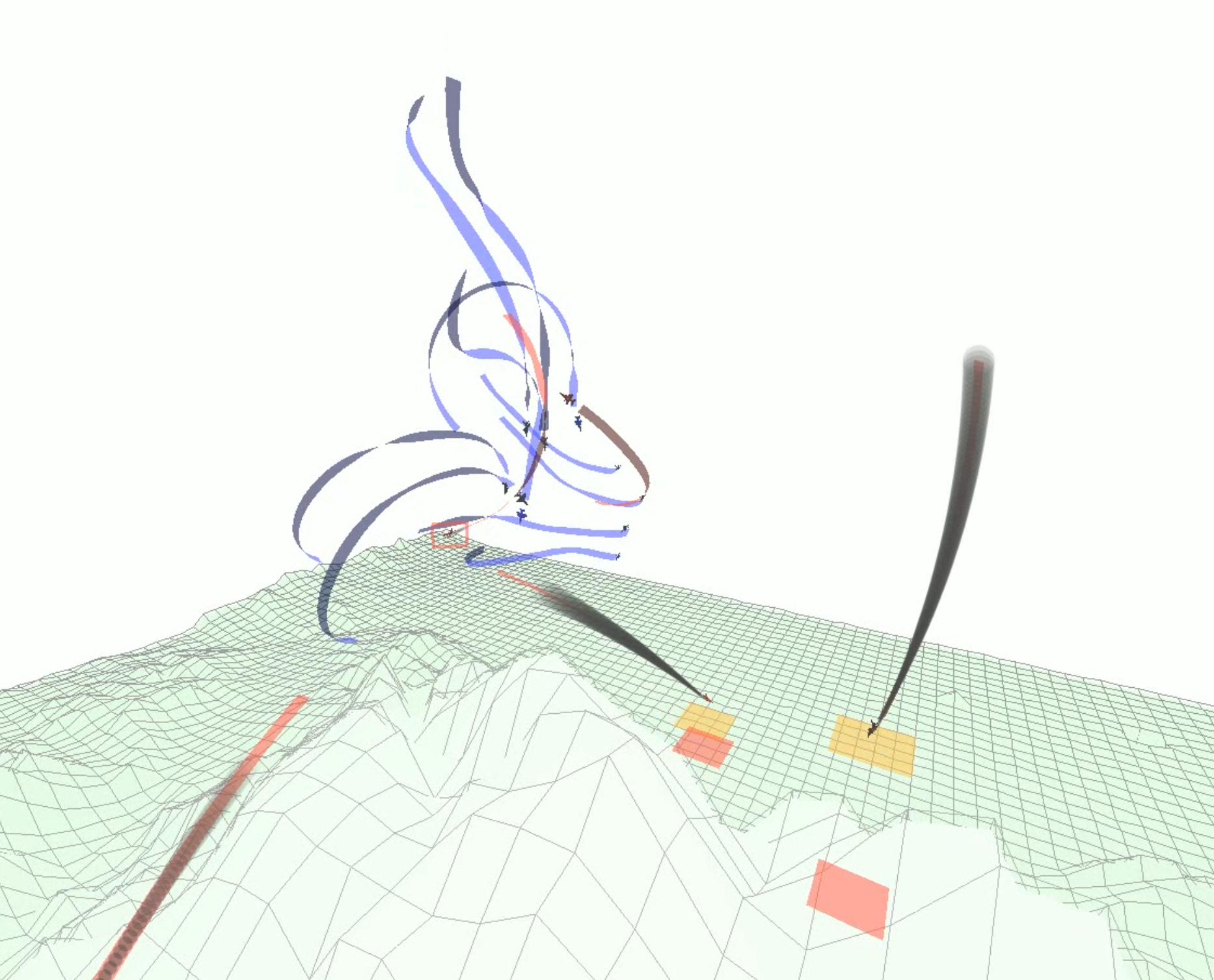}
\caption{Screenshot from 10v10 video showing red rectangles indicating an aircraft is in danger of being captured.  Once captured, an explosion is indicated, the aircraft loses all thrust, and smoke is emitted by the aircraft until it reaches the ground.  As the aircraft approach a minimum safe altitude known as the hard deck (1000 ft above the maximum terrain height) an animated yellow and red square under the aircraft indicate that the aircraft is receiving a penalty for being too close to the ground and is attempting to pull up in response.}
\label{sample_10v10}
\end{figure}

\section{Results}

In Figure \ref{plot_1v1_trajectory}, results are shown for a typical 1 versus 1 (1v1) encounter.  As blue has a performance advantage, it is able to maneuver more effectively and is able to capture the red aircraft.  Figure \ref{plot_1v1_actions} shows the actions selected by the blue aircraft during this run, while Figure \ref{plot_1v1_dynamics} shows the values of the pseudo-6DOF state variables during the run.

The $P_{\text{win}}$ of the blue team for all experiments is shown in Table \ref{blue_team}.  This is an indicator that the algorithm is functioning correctly as the blue team was given an advantage in the selection of actions and in aircraft dynamics.  Better dynamics allows the aircraft to maneuver into an offensive position more readily, leading to an expected high $P_{\text{win}}$.  Also as expected as the airspace volume becomes more crowded and complex due to the increase in team size, the probability of survivability $P_s$ tends to decrease.

\begin{table}[h]
\caption{Probability of win $P_{win}$ and Probability of survivability $P_s$ of blue team as team size increases}
\label{blue_team}
\begin{center}
\begin{tabular}{ |c|c|c| } 
 \hline
 Team Size & $P_{\text{win}}$  & $P_{s}$ \\
 \hline
 1v1 & 100\% & 100\% \\
 2v2 & 100\% & 100\% \\
 3v3 & 100\% & 100\% \\
 4v4 & 100\% & 100\% \\
 10v10 & 100\% & 99\% \\
 100v100 & 100\% & 97\% \\
 \hline
\end{tabular}
\label{team_size_results}
\end{center}
%\vspace{-15pt}
\end{table}

The amount of processing time required to formulate and solve the MDP for each agent at each timestep is summarized in Table \ref{processing_time}.  Processing was performed on a laptop with an Intel i9-8950HK CPU at 2.90 GHz.  While the code is written in Python, it does take advantage of the Numba and Numpy Python libraries that are used to perfom optimized computation loops in C.  Additionally, the underlying LLVM library may allow some Numba optimized code to take advantage of SIMD instruction in the CPU.  No GPU acceleration is used.

\begin{table}[h]
\vspace{5pt}
\caption{Processing time required for each agent on red or blue team as team size increases}
\label{processing_time}
\begin{center}
\begin{tabular}{ |c|c| } 
 \hline
 Team Size & Mean (ms) \\
 \hline
 1v1 & 2.26 \\ 
 2v2 & 2.50 \\ 
 3v3 & 2.70 \\
 4v4 & 3.16 \\
 10v10 & 5.55  \\ 
 100v100 & 27.59 \\
 \hline
\end{tabular}
\end{center}
\end{table}

Videos of example runs of 1v1, 2v2, 3v3, 4v4, and 10v10 are available for viewing are provided in Table \ref{video_links}.  Note that the size of the aircraft is exaggerated by a factor of 3 for improved visibility in the video.

\begin{table}[h]
\caption{Links to videos}
\label{video_links}
\begin{center}
\begin{tabular}{ |c|c| } 
 \hline
 Team Size & URL \\
 \hline
    1v1   & \url{https://youtu.be/zGWXxtJUwk8} \\
    2v2   & \url{https://youtu.be/Q9O50cqpVtA} \\
    3v3   & \url{https://youtu.be/6Zok4sj43C4}  \\
    4v4   & \url{https://youtu.be/qhI6av3oJN4} \\
    10v10 & \url{https://youtu.be/6twTWNRurwo} \\
 \hline
\end{tabular}
\end{center}
\end{table}

\section{Conclusion}

We have presented an efficient problem formulation for pursuit/evasion problems that scales to large numbers of teams (100v100) while remaining computationally efficient.  This method formulates the problem as a Markov Decision Process (MDP) and uses a recently proposed approach in \cite{bertram2019} to efficiently solve the MDP and is suitable for embedded systems commonly found on aircraft.  The use of ``risk wells" to represent the potential future actions of friendly and opposing aircraft  allows the problem to remain tractable even as the number of aircraft per team increases.

For future work, we plan to explore how to incorporate mutual support, combat tactics, and multiagent cooperation to increase the effectiveness of the teams.  This should be a rich area to explore with ample problems to examine.
We also plan on extending the aircraft model used here to a higher fidelity model to test the algorithm in different areas of the flight envelope.

\bibliographystyle{unsrt}
\bibliography{refs}

\end{document}